\documentclass[letterpaper, 10 pt, conference]{ieeeconf}  

\IEEEoverridecommandlockouts                              

\usepackage{graphicx}
\usepackage[maxbibnames=1, sorting=none, style=numeric-comp, doi=false, url=false]{biblatex}
\usepackage{algorithm}
\usepackage{algpseudocode}
\usepackage{tabularx}

\algrenewcommand\algorithmicindent{1.0em}%

\title{\LARGE \bf
Vision-based FDM Printing for Fabricating Airtight Soft Actuators
}

\author{Yijia Wu,$^{1, \dagger}$ Zilin Dai,$^{1, 2, \dagger}$ Haotian Liu,$^{1, 2}$ Lehong Wang,$^{1, 2}$ and Markus P. Nemitz$^{1, 3, 4}$
\thanks{$^{\dagger}$These authors contribute equally to this work}%
\thanks{This work was supported by the National Science
Foundation under CAREER Grant No. 2237506.}
\thanks{$^{1}$Department of Robotics Engineering, Worcester Polytechnic Institute, 100 Institute Road, Worcester, 01609 MA, USA
        {\tt\small mnemitz@wpi.edu}}%
\thanks{$^{2}$Department of Computer Science, Worcester Polytechnic Institute, 100 Institute Road, Worcester, 01609 MA, USA}%
\thanks{$^{2}$Department of Mechanical and Materials Engineering, Worcester Polytechnic Institute, 100 Institute Road, Worcester, 01609 MA, USA}%
\thanks{$^{3}$Department of Electrical and Computer Engineering, Worcester Polytechnic Institute, 100 Institute Road, Worcester, 01609 MA, USA}%
\thanks{Project code and more details will be available in  \url{https://github.com/roboticmaterialsgroup/closed-loop-printing}}
}
\begin{document}

\maketitle
\thispagestyle{empty}
\pagestyle{empty}

\begin{abstract}

Pneumatic soft robots are typically fabricated by molding, a manual fabrication process that requires skilled labor. Additive manufacturing has the potential to break this limitation and speed up the fabrication process but struggles with consistently producing high-quality prints. We propose a low-cost approach to improve the print quality of desktop fused deposition modeling by adding a webcam to the printer to monitor the printing process and detect and correct defects such as holes or gaps. We demonstrate that our approach improves the air-tightness of printed pneumatic actuators while reducing the need for fine-tuning printing parameters. Our approach presents a new option for robustly fabricating airtight, soft robotic actuators.

\end{abstract}

\section{INTRODUCTION}

Pneumatic actuators are the most popular soft actuators because of their low cost and simplicity \cite{soft_robot_control}. Currently, most pneumatic soft robots are manufactured by silicone molding \cite{soft_robot_manufacturing_review}, requiring training and time-consuming fabrication processes for complex structures. Additive manufacturing provides the advantages of reducing fabrication time and building structures that cannot be implemented by silicone molding \cite{embedded_3d_printing, 3d_printing_soft_robot_review, 3d_printing_bending_actuator}. However, most of the existing additive manufacturing methods used for the fabrication of soft structures, such as stereolithography, inkjet printing, or laser powder bed fusion, require expensive printers and consumables including stock materials, limiting adoption to specialized laboratories. 

In comparison, fused deposition modeling (FDM) is more cost-efficient and is the most commonly used additive manufacturing method. However, the fabrication of high-quality, air-tight soft robotic systems presents significant challenges with FDM due to the inherent slippage and buckling behavior of thermoplastic polyurethanes (TPU) \cite{TPU_printing_slippage} and the material-specific inconsistencies in extrusion when utilizing commercial FDM printers \cite{FFF_printing_review}. Although TPU filaments with low shore hardness are particularly interesting for the fabrication of soft robots (e.g., Filaflex from Recreus, 60A), severe leakage problem limits their popularity.

\begin{figure}[h]
\centering
    \includegraphics[width=0.9\linewidth]{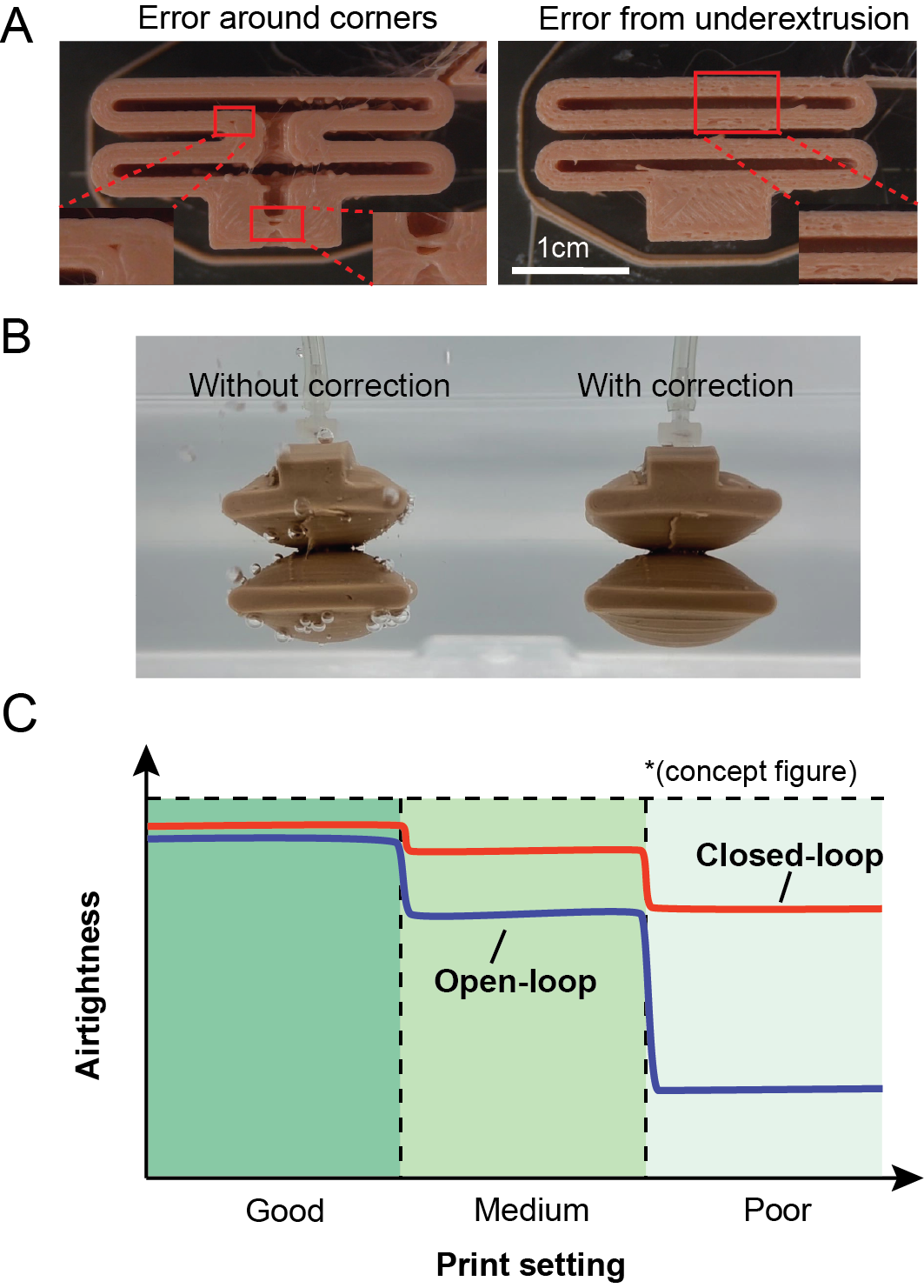}
    \caption{\textbf{Overview of our closed-loop, vision-based printing strategy for FDM.} (A) Printing defects that could lead to leakage but cannot be fixed with existing methods. (B) Photos of using the same setup and parameters to print a bellow actuator. When the correction subsystem is not in use, the bellow actuator leaks (left), while the bellow actuator printed with the correction subsystem is fully airtight (right). (C) Concept figure: our closed-loop printing strategy has the potential to improve the airtightness of soft systems in comparison to open-loop printing strategies; we detect and correct defects during the printing process.}
    \label{fig:intro}
\end{figure}

Researchers have proposed different solutions to mitigate the leaking problem in FDM-printed structures. 1) Dry filament before printing to avoid print defects through vaporized water at the print nozzle \cite{dehydrate_filament}; 2) Increase shell thickness \cite{air_connector_4_layer}; 3) Print designs with a single, continuous toolpath \cite{desktop_fabrication_scirob}; 4) Reduce print speed and over-extrude via extrusion multiplier and overlap ratio to fill potential gaps (print errors) organically \cite{penunet_ninjaflex_large_ext_multiplier_slow, helical_actuator_filaflex_slow, printable_robot_overextrude}. However, these approaches yield limited success in the fabrication with filaments of low shore hardness ($< 80A$) and are insufficient for mitigating stochastic anomalies, such as print defects that result from batch-to-batch variations in filament quality (\textbf{Figure~\ref{fig:intro}A}).

The implementation of a closed-loop FDM printing system for real-time detection and rectification of holes and gaps during the thermoplastic polyurethane fabrication process has the potential to yield impermeable structures (\textbf{Figure~\ref{fig:intro}BC}). While prior studies have successfully utilized camera-based systems for the identification of prevalent print defects like stringing \cite{stringing_detection}, layer-shift, deformed infill, and warping \cite{open_source_layerwise}, a majority of these defects either remain irreparable or necessitate manual intervention for correction. For amendable defects, existing techniques adjust print parameters to rectify issues such as over-extrusion, under-extrusion, and warp deformation in subsequent printing processes \cite{insitu_correction_nature, warp_deformation_correction}. However, these methods lack the capability for real-time corrections during the print cycle \cite{CMU_closed-loop}. The majority of existing closed-loop printing systems focus on rigid filaments, making some design approaches non-transferable due to the unique material properties of soft filaments.

In this paper, we introduce a layer-wise monitoring and control framework to enhance the reliability of fabricating airtight soft systems. Following the completion of each layer, a camera captures the current print status. Should defects be identified, these errors are rectified before progressing to the subsequent layer. The primary contributions of our research are:
\begin{enumerate}
  \item Design of a closed-loop FDM printing system optimized for real-time detection and remediation of airtightness-related defects.
  \item Formulation of a software architecture capable of executing layer-wise defect detection and correction through whole-layer ironing techniques.
  \item Empirical validation of our integrated hardware and software architecture by enhancing the airtightness characteristics of fluidic linear actuators under specific print parameters.
\end{enumerate}

\begin{figure}[t]
    \centering
    \includegraphics[width = 1\linewidth]{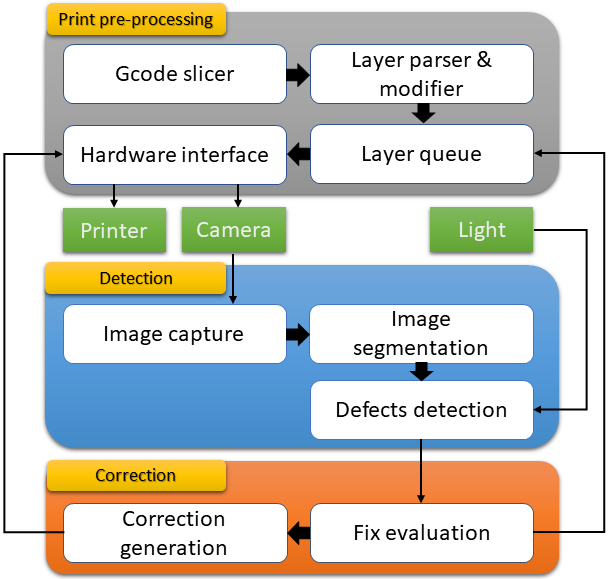}
    \caption{\textbf{Block diagram of our closed-loop FDM printing system.} Green blocks represent hardware interfacing with software subsystems. \textbf{Print pre-processing:} processes the original G-code into a layer-by-layer format, embedding movement commands for image capture, and sends comments to the printer. \textbf{Detection:} identifies defective areas in the current layer through real-time image analysis. \textbf{Correction:} rectifies printing errors within the current layer.}
    \label{fig:block_diagram}
\end{figure}

\section{SYSTEM DESIGN}

Our closed-loop printing system for fabricating airtight soft structures encompasses three primary subsystems: A) Print pre-processing, B) Detection, and C) Correction (\textbf{Figure~\ref{fig:block_diagram}}). These modules sequentially issue commands to the printer, detect anomalies during fabrication, assess the identified defects, formulate corrective strategies, and execute the devised remediations.

\begin{algorithm}
\caption{Defect Detection and Correction}
\begin{algorithmic}
    
\State $Hardware Setup $($SerialPort$, $camera$)
\State $AllLayerwiseGcode \leftarrow LayerParsing(GcodePath)$
\State $LayerModification (AllLayerwiseGcode)$
\For{$\forall layer \in AllLayerwiseGcode$}
    \State $SendGcode(SerialPort, layer)$
    \State $image \leftarrow ImageCapture(camera)$
    \State $defect \leftarrow DetectDefect(image, layer)$
    \State $d \leftarrow number\ of\ defect$
    \If{$d<threshold$} 
        \State \textbf{pass}
    \Else
        \State $CGcode \leftarrow GenerateCorrectionGcode (layer)$
        \State $SendGcode(SerialPort, CGcode)$
    \EndIf
\EndFor

\end{algorithmic}
\end{algorithm}

\subsection{Print pre-processing}
The print pre-processing subsystem serves as the preparatory phase for the ensuing closed-loop detection and correction activities. It manages hardware communication among the printer, camera, and computer, and also handles G-code manipulation. Standard FDM slicers convert a 3D model into layer-wise toolpaths, encapsulating all printing instructions within a G-code file. This includes directives such as nozzle movement and flow rate adjustment. Upon receipt of a G-code file, the printer executes these instructions sequentially. We employ \textit{Printrun}, an open-source library, for direct computer-based control of the FDM printer.

\begin{figure*}[h]
    \centering
    \includegraphics[width = 0.95\linewidth, height = 3cm]{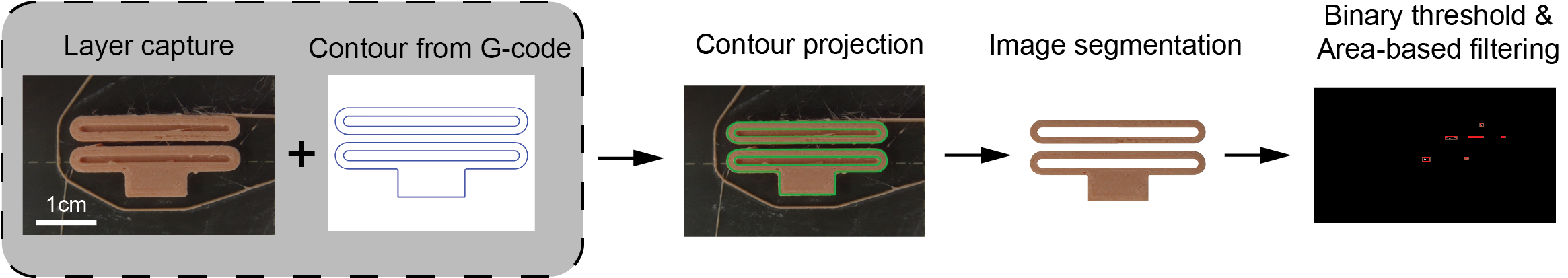}
    \caption{\textbf{Defect detection pipeline.} For each layer, contours are extracted from the G-Code and mapped onto the captured image. Utilizing these projected boundaries, the most recently printed layer is segmented from the overall image. Subsequently, areas of low intensity are filtered from this segmented portion. Regions falling within a pre-defined size range are classified as defects.
    }
    \label{fig:detection}
\end{figure*}


During initialization, the print pre-processing subsystem sets up connections among the computer, camera, and printer, and prepares the G-code files for subsequent execution. It parses the monolithic G-code files from the slicer into distinct layer-wise files, embedding commands for image capture. In the actual printing phase, a layer-wise print-detect-correct routine is carried out (\textbf{Algorithm 1}). This subsystem is tasked with dispatching the next G-code file from the instruction queue to the printer for execution.



\subsection{Detection}

The detection subsystem assesses the quality of the most recently printed layer using image-based evaluation (\textbf{Figure~\ref{fig:detection}}). Open Source Computer Vision Library (OpenCV\cite{opencv_library}) facilitates image capture and executes foundational image processing functions.

\subsubsection{\textbf{Image capturing}}

The initial step in detection involves capturing high-resolution images to facilitate defect identification. Our objective at this juncture is to detect holes and gaps, which are often indicative of potential leakage. Originating from inconsistent extrusion, these defects typically have widths approximating the printer nozzle's diameter (0.4 $mm$ in our setup). To enhance defect visibility in images, the camera is mounted on the print head, ensuring a small and constant distance to the printed layer. This setup negates the need for focal adjustments when layer heights vary, consistently capturing a large defect area within each image.



\begin{figure}[h]
    \centering
    \includegraphics[width=0.95\linewidth]{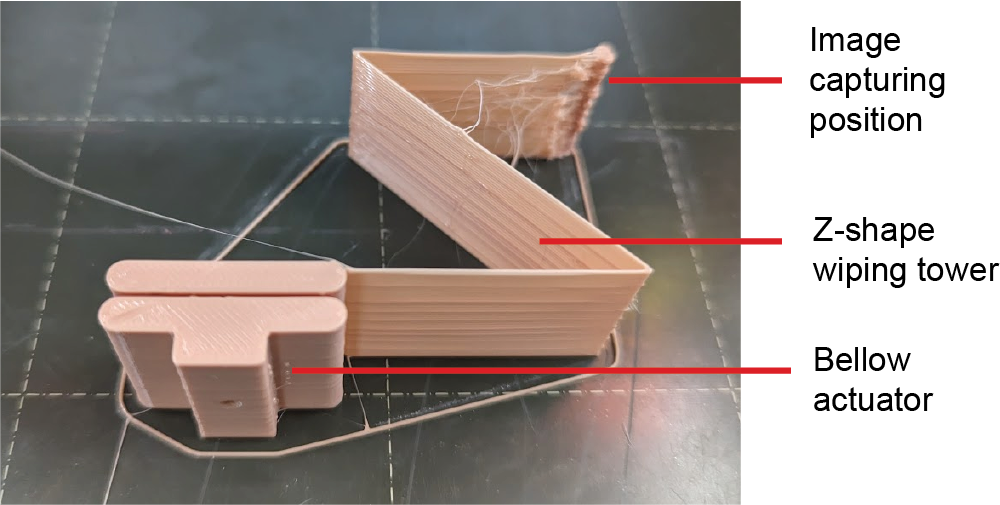}
    \caption{\textbf{Z-Shape wiping tower for nozzle cleaning.} The Z-shaped structure serves as a transition path for the nozzle, moving from the image capture location to the subsequent printing position. This pathway allows the nozzle to wipe off any oozed filament, ensuring optimal extrusion conditions before resuming printing.}
    \label{fig:z_pattern}
\end{figure}

Upon completing a layer, the camera is repositioned to capture the entire layer, necessitating that the nozzle moves outside the original printing zone. A brief pause is required to stabilize the print head before capturing the image, causing some filament oozing and subsequent under-extrusion. To mitigate this issue without compromising image quality or printing accuracy, we implemented a Z-shaped structure (\textbf{Figure~\ref{fig:z_pattern}}), situated at the object's upper-right corner, to act as a wiping mechanism. After each image capture, the nozzle prints this Z-shaped structure before resuming object printing, thereby negating the effects of filament oozing. The wiping structure matches the object's height and has a width equal to the nozzle diameter.

\subsubsection{\textbf{Image segmentation}}
Upon obtaining a high-quality image, the initial step of detection involves cropping the region corresponding to the most recently printed layer from the raw image. This ensures that detected errors are specific to the current layer.

Initially, various segmentation techniques were explored, but differentiating between current and preceding layers proved challenging from a top-down perspective due to the lack of distinct visual cues. The ultimate segmentation method employs contours projected from the G-code to demarcate boundaries. The external perimeters in the G-code are extracted as the contours. They graphically define the outer boundary of each print layer (see \textbf{Figure~\ref{fig:detection} Contour from G-code}). These contours provide 3D positional data points for the external perimeter of the relevant layer within the printer frame, acting as a reference outline for the target shape. The contour points can be mapped from the 3D printer frame onto the 2D image plane using the \texttt{cv.ProjectPoints} function, provided the relative transformation between the printer and camera frames is known (see \textbf{Figure~\ref{fig:detection} Contour Projection}).

In our setup, we consider three coordinate frames: \(\textbf{\textit{C}}\) for the camera, \(\textbf{\textit{N}}\) for the nozzle, and \(\textit{\textbf{B}}\) for the print bed. Given that the camera is affixed to the print head, the transformation matrix \( T_\textbf{\textit{CN}} \) can be determined through extrinsic calibration. Similarly, the transformation matrix $T_\textbf{\textit{NB}}$ is known since the nozzle position is governed by the G-code. Consequently, the transformation matrix from the printer frame to the camera frame can be formulated as follows:


\begin{equation}
    T_{\mathbf{\textbf{CB}}} = T_{\textit{\textbf{CN}}} \cdot T_{\mathbf{\textbf{NB}}}
\end{equation}


Ultimately, the most recent printed layer is isolated from the image by successively converting each contour to a mask and executing bitwise XOR operations between these masks and an initialized zero-value mask (see \textbf{Figure~\ref{fig:detection} Image Segmentation}).


\subsubsection{\textbf{Defect detection}}

The final component involves a defect identification algorithm applied to the segmented images, consisting of two key steps: binary thresholding and area-based filtering.



Our lighting setup ensures that under-extruded regions are darker than properly extruded regions because of the lack of lighting, so we can apply a binary threshold to segment grayscale images to isolate potential defect areas. 
Utilizing the \texttt{cv.connectedComponentsWithStats} function, we obtain size metrics for each connected dark region (depicted as white in \textbf{Figure~\ref{fig:detection}}). Areas falling outside of pre-defined minimum and maximum size thresholds—likely representing noise or other types of print defects—are eliminated. The count of remaining areas serves as the defect tally. The minimum and maximum size thresholds should be chosen according to the potential defect size and spatial resolution of the camera. The potential defect size is proportional to the nozzle size. We suggest a range between 1 and 20 times the nozzle diameter. The spatial resolution, defined by the distance each pixel represents, is influenced by the image resolution, camera focal length, and distance from the camera to the printed layer. Given a representation where 1 mm is represented by 35.8 pixels and a nozzle size of 0.4 mm, we chose 15 and 300 as the minimum and maximum size thresholds, respectively.

\subsection{Correction subsystem}
The correction subsystem interprets defect data from the detection subsystem to generate the appropriate corrective G-code file.


\subsubsection{\textbf{Fix evaluation}}
For each layer, the number of defects identified by the detection subsystem is compared against an empirically-tuned threshold to decide whether a correction is warranted. This threshold balances correction time and the likelihood of achieving an airtight structure.

\subsubsection{\textbf{Whole-layer ironing correction}}
If the number of filtered defects exceeds the threshold, the system modifies the original G-code for the current layer, reducing the extrusion rate and movement speed (20\% and 60\% in our setup).  With the print head height unchanged, the heated nozzle would iron the layer while minimally extruding material. This modified G-code is saved and sent to the printer for execution. After correction, another round of detection could be used to re-evaluate whether this layer requires further correction.

The whole-layer ironing process simplifies the camera extrinsic calibration required for fine-scale fabrication by minimally extruding material at reduced speeds. WInitially, we aimed to map detected defects to precise printer coordinates to minimize correction time. However, this method was less effective due to inaccuracies in camera calibration and the diminutive scale of the prints. The instability of extrusion in small areas hindered direct defect filling. As an alternative, we explored ironing each layer during printing to ensure a defect-free production process. Through empirical testing, we found that ironing every layer led to excessive material overlay, causing actuator chambers to adhere to one another and compromising the intended inflatable structure.

\begin{figure}[h]
    \centering
    \includegraphics[width=\linewidth]{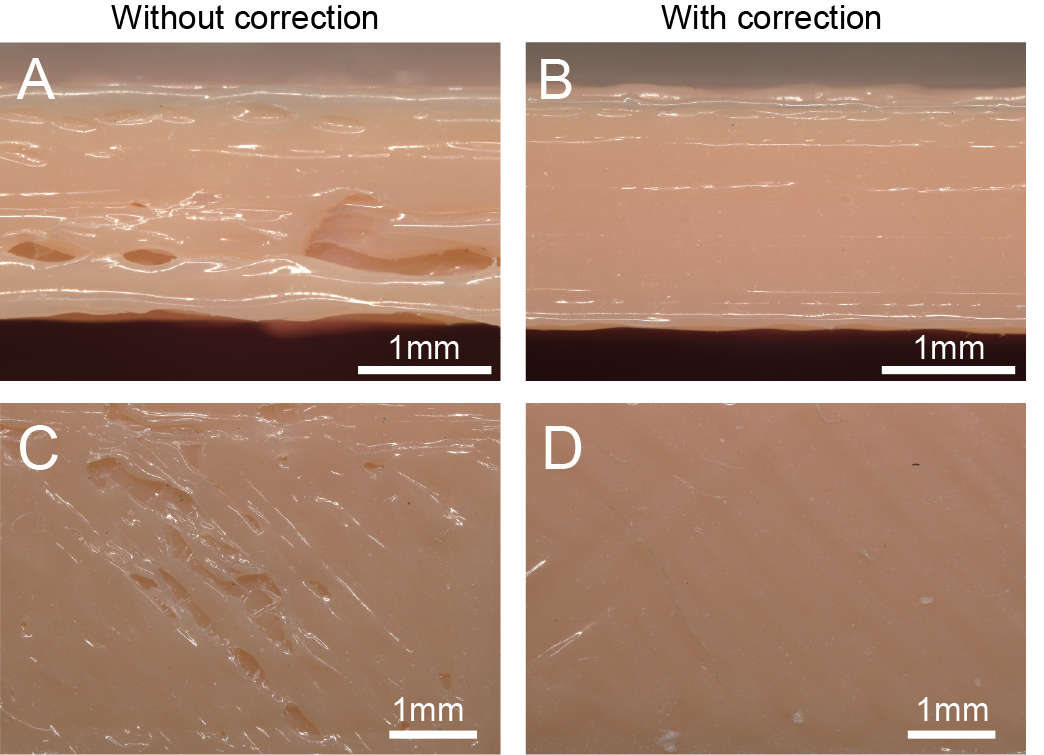}
    \caption{\textbf{Microscope image of uncorrected and corrected layers.} (A, C) Uncorrected holes within the structure, particularly in thin-walled sections, compromise the airtightness of the structure. (B, D) The correction process employs targeted melting of adjacent materials and supplemental extrusion to effectively seal the holes.}
    \label{fig:correction}
\end{figure}

In addition to filling material voids caused by random errors, rapid correction maintains nozzle heat continuity, mitigating risks of stringing and material hardening that could compromise the print. The Z-shaped wiping tower further stabilizes extrusion flow before commencing the subsequent layer. The outlined correction strategy effectively mitigates under-extrusion arising from mechanical uncertainties, offering a real-time, robust, and adaptive solution for achieving airtight soft material structures (\textbf{Figure~\ref{fig:correction}}).

\section{EXPERIMENTS AND RESULTS}

\subsection{System setup}
\begin{figure}[h]
    \centering
    \includegraphics[width=0.9\linewidth]{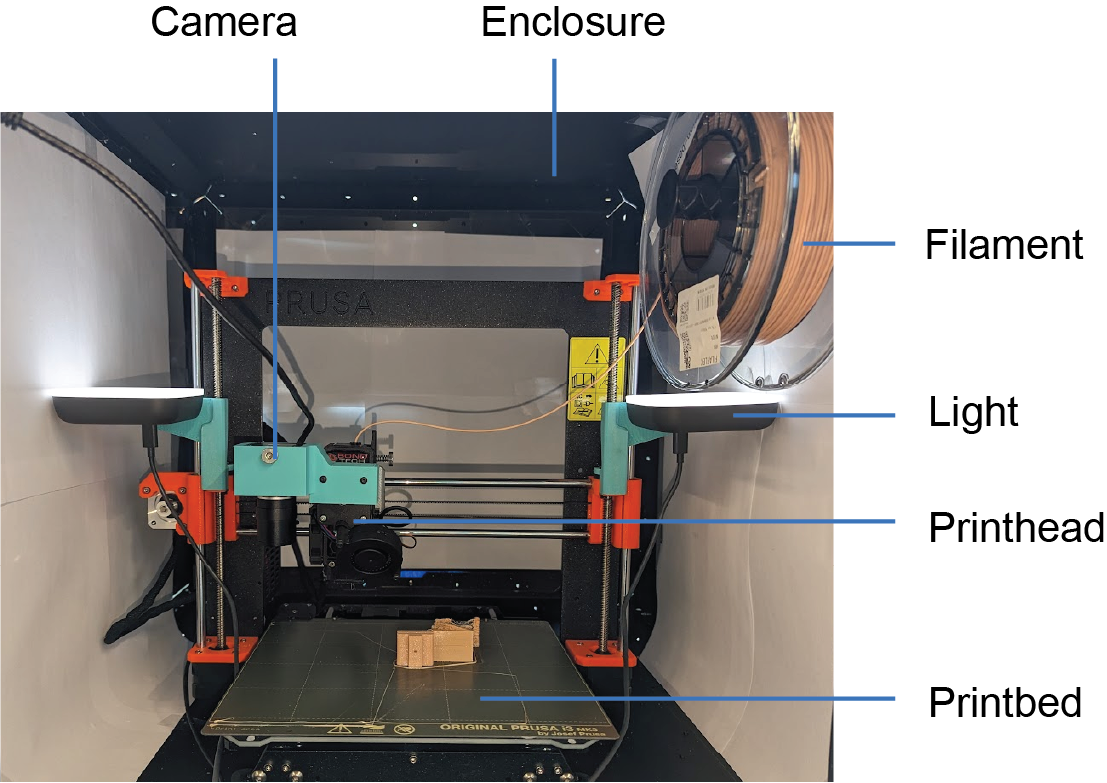}
    \caption{Hardware setup of the vision-based system. }
    \label{fig:system_setup}
\end{figure}

Our system utilizes a Prusa MK3S desktop FDM printer, housed within an Original Prusa Enclosure. The enclosure is modified with external black cardboard to block ambient light and internal white diffusion boards to disperse light from two Logitech Litra Glow sources. These light sources, along with an ELP IMX317USB 4K Webcam, are affixed to the z-axis slider and print head, enabling vertical movement in sync with the printing process (\textbf{Figure~\ref{fig:system_setup}}).



\subsection{Experiment design}
To evaluate the performance of our closed-loop system, we designed an experiment that compares the airtightness of bellows actuators fabricated under both open-loop and closed-loop control architectures.We selected Filaflex 60A by Recreus as our test material, recognizing it as the commercially available TPU filament with the lowest Shore hardness we are aware of. Initially, we based our print settings on the "GENERAL FLEX" profile in Prusa Slicer software, later fine-tuning these parameters within the ranges recommended by the manufacturer. Our evaluation comprises three distinct sets of print parameters (Table 1), each selected based on empirical data, to generate actuators with differentiated airtightness profiles. Increasing the layer height results in wider gaps between printed lines, elevating the risk of under-extrusion. This strategy allows for a multi-faceted assessment of the closed-loop system's efficacy under varying fabrication conditions.

\begin{table}[h]
\centering
\normalsize
\caption{\normalsize Print parameters}
\label{table: specific_parameter}
\begin{tabular}{ l|c|c|c } 
   & \textbf{Good} & \textbf{Medium} & \textbf{Poor} \\
  \hline
 \textbf{Nozzle temperature} $(^\circ C)$ & 220 &220 & 230 \\
 \textbf{Layer height} (mm) & 0.1 & 0.2 & 0.3 \\
\end{tabular}
\end{table}

The airtightness of the bellows actuators was measured through a leak rate evaluation, where each actuator was immersed in a water reservoir and exposed to a steady pressure of 100 kPa, using an AFR2000 pressure regulator to manage the airflow from a compressor. A burette, initially filled with water, is placed on top of the actuator to capture all escaped air with the help of a funnel. Leak rates were determined by dividing the volume of air escaped due to leakage by the corresponding elapsed time. To ensure statistical validity, five actuators were fabricated and tested under each set of print parameters.

\begin{figure}[h]
    \centering
    \includegraphics[width=\linewidth, trim={2cm 0.5cm 3cm 2cm},clip]{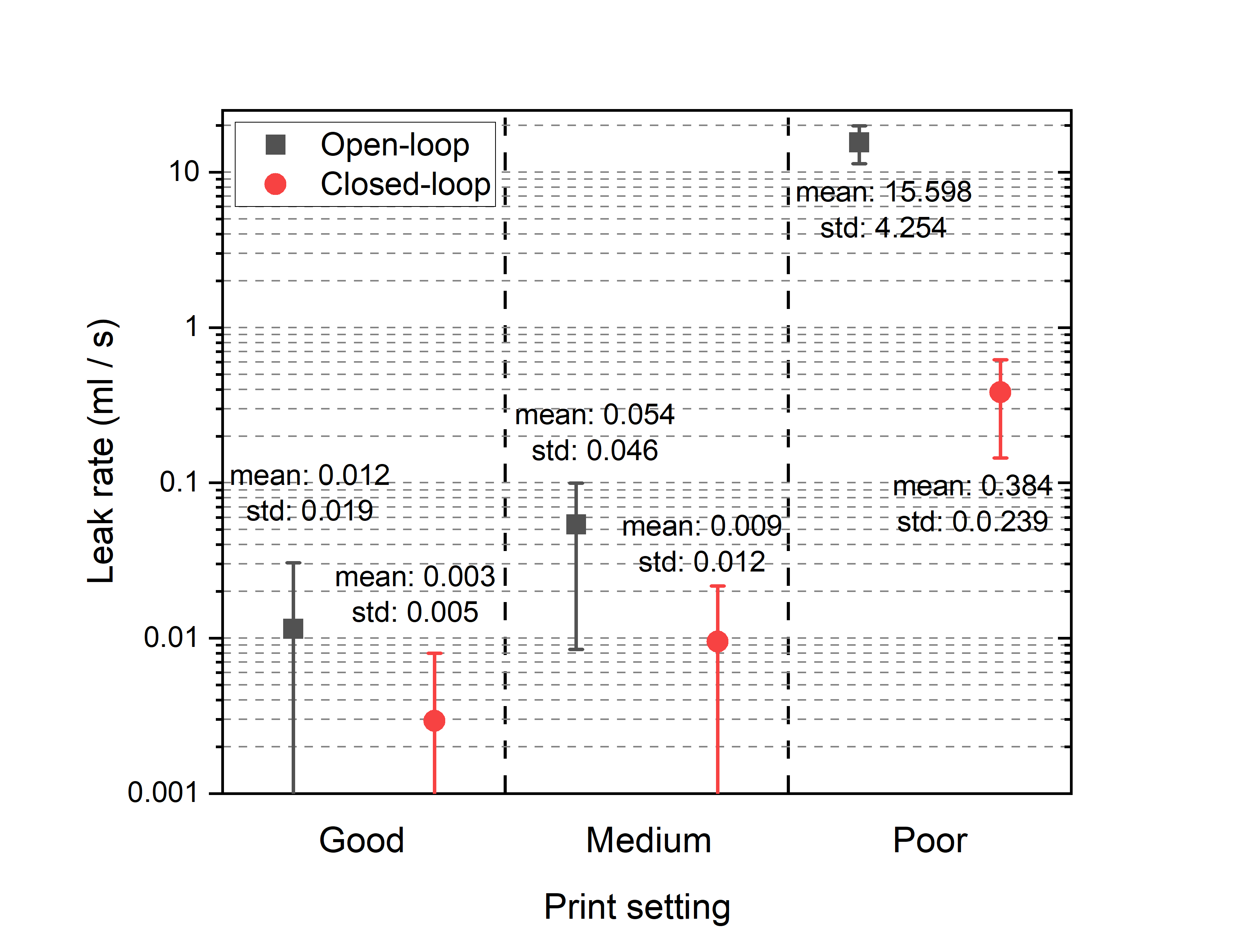}
    \caption{\textbf{Effect of the closed-loop printing system on airtightness.} The average and standard deviation of leak rate with three different sets of printing parameters (Table~\ref{table: specific_parameter}) across N = 5 samples are compared. The closed-loop system reduces the leak rate by 75.5\%, 82.4\%, and 97.5\% for the good, medium, and poor settings.}
    \label{fig:fix_evaluation}
\end{figure}

\subsection{Results}

Leak test results demonstrated that the closed-loop system enhances the airtightness of FDM-printed soft actuators under diverse printing scenarios (\textbf{Figure~\ref{fig:fix_evaluation}}). Notably, while improvements were observed in actuators printed with suboptimal parameters, residual leakage persisted, attributable to the inherent similarities between the corrective and initial printing processes, which are equally affected by the quality of the printing parameters.

A bulge line could be observed for every corrected layer because of the additional material. The average height of the bellow actuators and standard deviation are 23.92 $mm$ and 0.06 $mm$ for the open-loop system, and 24.01 $mm$ and 0.09 $mm$ for the closed-loop system.

By adjusting the ironing speed to 60\% of the standard printing speed, each layer requiring correction necessitates an additional 60\% printing time. In our experiments, between 13.8\% and 67.4\% of layers required correction, resulting in an 8.3\% to 40.4\% increase in total print time for the closed-loop system.

\section{DISCUSSION}


Our methodology enhances the airtightness of pneumatic soft systems, enabling generally applied parameters to match the performance of meticulously optimized parameters by expert users. Nevertheless, the efficacy of our system depends on certain assumptions inherent in its design, which, if not met, could limit its applicability.

\begin{enumerate}
\item The evaluation of the print is based solely on its airtightness, not on its dimensional accuracy.
\item The user is able to fabricate functional parts with their printer, and our approach is implemented to improve the airtightness of their prints.
\item Our approach only addresses random defects that are mainly caused by uncertainty while extruding elastic materials.
\end{enumerate}

Our approach is specifically tailored for adept users of desktop FDM printers who aim to enhance the air-tightness of their soft systems, bypassing the arduous task of parameter fine-tuning and avoiding the unpredictability of under-extrusion issues. This strategy presumes that the user's current printing setup and G-code are capable of producing a structure that is predominantly air-tight, capable of maintaining internal pressure and not deflating rapidly upon pressure release.

Our investigations revealed that the term "fixing" might suggest an overly optimistic capability of our approach; our method cannot rectify arbitrarily wrong print parameters. Experimental evidence instead indicates that our method enhances the quality of existing prints rather than rescuing fundamentally flawed ones. Hence, preliminary parameter optimization remains essential for the effectiveness of our proposed correction technique. 


Our correction strategy involves ironing entire layers, which can lead to unintended alterations in surface roughness, dimensional accuracy, and mechanical properties such as elasticity and compliance due to the deposition of extra material. This method is particularly apt for soft pneumatic actuators with minimal demands on precision or aesthetics, especially in phases where airtightness is prioritized. For components with stringent requirements on these attributes, the adoption of more sophisticated correction techniques will be necessary.

In this study, our defect detection technique relies on a binary threshold, making filament color a significant factor in its effectiveness. The contrast between well-extruded and under-extruded areas may be too subtle for accurate detection with dark or transparent filaments. Enhancing detection capabilities through the integration of additional sensors, like depth and thermal cameras, could address this issue, albeit at a cost potentially exceeding that of the FDM printer itself. Machine learning approaches offer a promising alternative, capable of more robust defect identification across varied lighting conditions and mitigating color-related limitations.


While analyzing the images of each layer, we identified slicing errors that resulted in unexpected holes, which had not been consistently accounted for alongside random errors. Despite the slicer allocating additional width to peripheral areas and print parameters being adjusted for over-extrusion, residual holes persist that are not rectifiable with the original setting. This system can also analyze the performance of the slicer and can assist in the development of tool path planning for printing airtight structures. 

\section{CONCLUSION}
In this study, we introduce a closed-loop FDM printing technique capable of rectifying unforeseen holes and gaps (printing discrepancies) through layer-specific detection and amendment. Our experimental findings reveal the system's versatility across various printing parameters, evidenced by the production of a bellows actuator under three distinct printing conditions. The employment of a vision-based monitoring system alongside layer-specific G-code adjustments has demonstrated to be both cost-efficient and effective, offering an innovative method to enhance the airtightness of FDM-printed soft systems.

\section*{ACKNOWLEDGMENT}
We thank Ritwik Pandey for assisting with the microscopy imaging of the printed components, and Songlin Hou for enhancing the photographic quality.

\printbibliography

\end{document}